\documentclass[sigconf]{acmart}
\usepackage{multirow}
\AtBeginDocument{%
  }

\setcopyright{cc}
\setcctype{by}
\acmConference[CIKM '25]{Proceedings of the 34th ACM International
Conference on Information and Knowledge Management}{November 10--14,
2025}{Seoul, Republic of Korea}
\acmBooktitle{Proceedings of the 34th ACM International Conference on
Information and Knowledge Management (CIKM '25), November 10--14, 2025,
Seoul, Republic of Korea}
\setcopyright{none}
\renewcommand\footnotetextcopyrightpermission[1]{}
\begin{document}

\title{BALM-TSF: Balanced Multimodal Alignment for LLM-Based Time Series Forecasting}

\author{Shiqiao Zhou}
\orcid{0009-0003-3515-2864}
\affiliation{%
  \institution{University of Birmingham}
  \department{School of Computer Science}
  \city{Birmingham}
  \country{United Kingdom}}
\email{sxz363@student.bham.ac.uk}

\author{Holger Schöner}
\orcid{0009-0009-9341-538X}
\affiliation{%
  \institution{Siemens AG}
  \city{Munich}
  \country{Germany}}
\email{holger.schoener@siemens.com}

\author{Huanbo Lyu}
\orcid{0009-0007-7007-6333}
\affiliation{%
  \institution{University of Birmingham}
  \department{School of Computer Science}
  \city{Birmingham}
  \country{United Kingdom}}
\email{hxl099@student.bham.ac.uk}

\author{Edouard Fouché}
\orcid{0000-0003-0157-7648}
\affiliation{%
  \institution{Siemens AG}
  \city{Nuremberg}
  \country{Germany}}
\email{edouard.fouche@siemens.com}

\author{Shuo Wang}
\authornote{Corresponding author}
\orcid{0000-0003-1380-6428}
\affiliation{%
  \institution{University of Birmingham}
  \department{School of Computer Science}
  \city{Birmingham}
  \country{United Kingdom}}
\email{s.wang.2@bham.ac.uk}

\thanks{© {Shiqiao Zhou, Holger Schöner, Huanbo Lyu, Edouard Fouché, and Shuo Wang | ACM} {2025}. This is the author's version of the work. It is posted here for your personal use. Not for redistribution. The definitive Version of Record was published in {CIKM '25}, http://doi.org/10.1145/3746252.3761278.}
\renewcommand{\shortauthors}{Shiqiao Zhou, Holger Schöner, Huanbo Lyu, Edouard Fouché, and Shuo Wang}

\begin{abstract}
Time series forecasting is a long‐standing and highly challenging research topic. Recently, driven by the rise of large language models (LLMs), research has increasingly shifted from purely time series methods toward harnessing textual modalities to enhance forecasting performance. However, the vast discrepancy between text and temporal data often leads current multimodal architectures to over-emphasise one modality while neglecting the other, resulting in information loss that harms forecasting performance. To address this modality imbalance, we introduce BALM-TSF (Balanced Multimodal Alignment for LLM-Based Time Series Forecasting), a lightweight time series forecasting framework that maintains balance between the two modalities. Specifically, raw time series are processed by the time series encoder, while descriptive statistics of raw time series are fed to an LLM with learnable prompt, producing compact textual embeddings. To ensure balanced cross-modal context alignment of time series and textual embeddings, a simple yet effective scaling strategy combined with a contrastive objective then maps these textual embeddings into the latent space of the time series embeddings. Finally, the aligned textual semantic embeddings and time series embeddings are together integrated for forecasting. Extensive experiments on standard benchmarks show that, with minimal trainable parameters, BALM-TSF achieves state-of-the-art performance in both long‐term and few‐shot forecasting, confirming its ability to harness complementary information from text and time series. Code is available at \url{https://github.com/ShiqiaoZhou/BALM-TSF}.
\end{abstract}

\ccsdesc[500]{Information systems~Data mining}
\ccsdesc[500]{Applied computing~Forecasting}

\keywords{Time Series Forecasting; Multimodal Learning; Prompt Learning; Large Language Model}


\maketitle

\section{Introduction}
Time series forecasting is essential to a wide range of real-world applications, including demand prediction \cite{seyedan2020predictive}, energy consumption forecasting \cite{daut2017building}, and financial market analysis \cite{ding2015deep}. In practice, target variables in time series are often affected by numerous external factors. Achieving accurate forecasts therefore requires not only robust modeling techniques but also the integration of relevant domain knowledge. Recent studies have investigated the use of auxiliary information—either by incorporating it directly as additional input features or by employing it as external conditioning signals during the forecasting process \cite{wang2024timexer, li2020multimodal}.

Among various types of supplementary data, textual information often provides rich and complementary context that pure time series models may fail to capture. Advances in natural language processing (NLP) have made large language models (LLMs) powerful tools for extracting semantics from text sources such as news, reports, and weather bulletins. Pre-trained on massive web-scale corpora, LLMs such as GPT series \cite{radford2019language,brown2020language, achiam2023gpt}, LLaMA \cite{touvron2023llama}, and DeepSeek \cite{liu2024deepseek} are capable of encoding textual inputs into meaningful representations for downstream tasks. Motivated by this, an increasing number of time series forecasting models have begun to integrate LLMs to extract and utilize textual information, demonstrating promising improvements in forecasting accuracy \cite{jin2023time, liu2024timecma, hu2025context,zhou2023one, williams2024context}.

While LLM-based time series forecasting has shown great potential, it introduces a fundamental challenge: \textbf{modality imbalance}, which manifests in both semantic and distributional forms. \textbf{Semantic imbalance} arises from the fact that large language models (LLMs) are pre-trained primarily on textual data, 
\begin{figure}[h]
  \centering
  \includegraphics[width=\linewidth, trim = 6cm 2cm 7cm 1cm, clip]{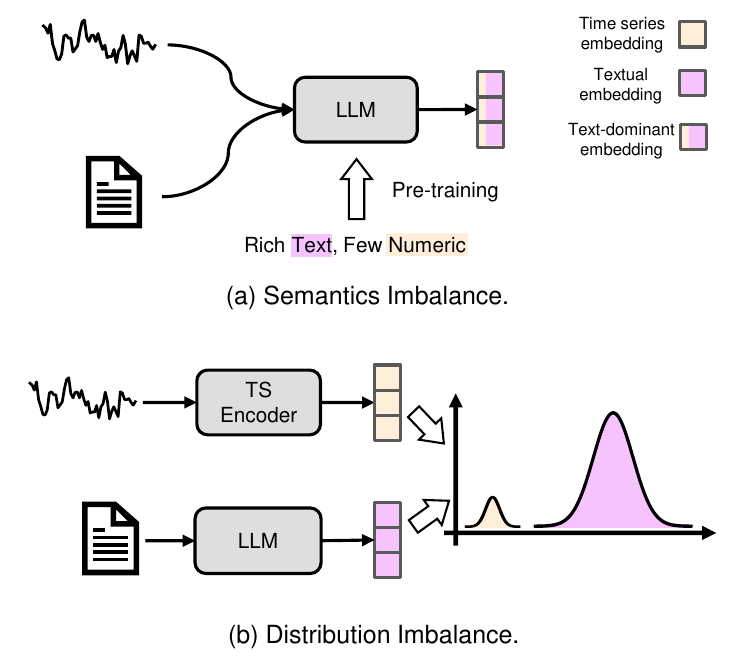}
  \caption{Modality imbalance issue in LLM-based time series forecasting.
(a) Semantic Imbalance: LLMs pretrained on rich text but sparse numeric data default to text patterns when handling time series inputs, resulting in “text-dominant” embeddings.
(b) Distributional Imbalance: Separately encoded time series and text embeddings often have mismatched value ranges, hindering effective downstream fusion.}
  \Description{}
  \label{MI}
\end{figure}
with minimal exposure to numerical patterns~\cite{lewkowycz2022solving}. In fact, numeric data such as time series center on dynamic trends and precise numeric changes, whereas text conveys static semantics and discrete events—creating a wide representational gap. Consequently, LLMs demonstrate weak numerical reasoning and struggle to capture the fine-grained quantitative fluctuations essential for accurate forecasting.

Furthermore, Modern LLM tokenizers often handle numbers in a suboptimal way. Instead of treating numbers as single units, tokenizers split them into subwords based on corpus statistics, producing inconsistent and sparse numeric representations~\cite{singh2024tokenization}. For instance, in the GPT-3 tokenizer, a frequently occurring small integer like “567” is mapped to a single token, whereas a slightly less common or larger number such as “568” is inconsistently split into multiple sub-word tokens, yielding discontinuous encodings~\cite{singh2024tokenization}. Such discontinuous encodings hinder models from inferring magnitude or place value, especially in long or floating-point numbers where each digit or decimal is tokenized separately ~\cite{schwartz2024numerologic}. As a result, LLMs perform poorly on reasoning tasks involving long or complex numbers ~\cite{yang2024number}. This results in coarse numeric encodings that limit LLMs’ ability to process continuous values such as time series.

Mainstream models such as Time-LLM~\cite{jin2023time} and UniTime~\cite{liu2024unitime} overlook this semantic imbalance. They integrate textual and temporal modalities by feeding both directly into the LLM. However, because LLMs are trained on text-dominant corpora rich in language but poor in numerical data, they treat time-series values as discrete tokens rather than continuous signals. Consequently, the attention layers overemphasize linguistic patterns and underplay true temporal dynamics, producing a “text-dominant” embedding that obscures the underlying time series information, as shown in Figure~\ref{MI}(a). This raises concerns regarding the effectiveness of LLMs in time series forecasting~\cite{tan2024language}.

Another category of models with a dual-branch structure appears to mitigate semantic imbalance issue. Recent works have adopted dual-encoder architectures, separately encoding text and time series followed by cross-modal alignment \cite{liu2024timecma, liu2025calf, wang2025tshtfaadvancingtimeseries}. While such designs mitigate semantic interference, they often face another issue: \textbf{distribution imbalance}. Typically, time series embeddings are normalized (e.g., via RevIN~\cite{kim2021reversible}) at beginning, which make their value range be dense to zero. As an illustrative example, after time series encoder, they might span from -13.12 to 14.12 with a near-zero mean of 0.00011 and low standard deviation of 0.79. In contrast, textual embeddings from LLMs—again as an example—can range from -36.00 to 45.50 with a mean of \(0.012\) and a higher standard deviation of 1.35, highlighting the necessity of cross-modal scaling. As shown in Figure~\ref{MI}(b), this results in substantial distributional mismatch. In further multimodal fusion, the higher-variance modality may dominate gradient updates, degrading alignment and harming forecasting performance. Although some models attempt complex alignment modules to mitigate this, they often introduce excessive computational overhead without clear gains \cite{liu2025calf, liu2024timecma}.

To address modality imbalance, we propose BALM-TSF (Balanced Multimodal Alignment for LLM-Based Time Series Forecasting), a simple yet effective framework that decouples the two modalities into separate branches. Instead of feeding raw time series into the LLM, we first summarize each series via descriptive statistics to form a compact textual prompt, then augment this prompt with a small set of learnable prompt tokens, activating the LLM to extract high-level semantic information without ever processing raw time series values directly. We then apply a two-step balanced multimodal alignment strategy: (1) Scaling, which resizes the textual embedding to match the variance of the time series embedding while compressing relevant tokens based on prediction length; (2) Semantic alignment, which uses a contrastive loss to ensure that the text ual and temporal representations of the same instance are closely aligned, while unrelated ones remain distant. Finally, the aligned embeddings are concatenated and passed to a shared forecasting head. Through its simple yet effective design, our model alleviates the modality imbalance between temporal and textual inputs in LLM-based forecasting and outperforms a wide range of state-of-the-art baselines in both long-term and few-shot prediction scenarios.

In summary, this work makes the following key contributions:%
\begin{itemize}
    \item We identify and analyze the modality imbalance issue inherent in LLM-based time series forecasting, where textual representations tend to dominate in semantics and value distribution of embedding.
    
    \item We present BALM-TSF, which addresses modality imbalance through a dual-branch design with learnable prompt and a simple yet effective balanced alignment strategy based on scaling and contrastive objectives.
    
    \item We show that BALM-TSF outperforms state-of-the-art baselines with significantly fewer parameters, establishing it as a lightweight and effective benchmark for multimodal time series forecasting.
\end{itemize}

\section{Related Work}
\subsection{Time Series Forecasting with LLM}
Time series forecasting models range from classical ARIMA \cite{box2015time} and LSTM \cite{hochreiter1997long} to Transformer-based architectures like PatchTST \cite{Yuqietal-2023-PatchTST} and iTransformer \cite{liu2023itransformer}. Recently, LLMs have shown potential to transform time series analysis. Thanks to advances in LLM architectures~\cite{radford2019language, brown2020language, achiam2023gpt, touvron2023llama, liu2024deepseek} and demonstrated multimodal capabilities, researchers have begun to adapt LLMs for time series forecasting~\cite{jin2024position}. Early work leveraged GPT-2’s transformer blocks and positional encodings to capture temporal patterns~\cite{jia2024gpt4mts, zhou2023one}. The subsequent work, Time-LLM~\cite{jin2023time}, reprogrammed LLM embeddings directly for the time series domain. UniTime~\cite{liu2024unitime} learned a unified representation across multiple datasets with diverse text inputs. More recent methods like FSCA \cite{hu2025context} jointly models time series and text using graph neural networks to capture cross-modal dependencies. TimeCMA \cite{liu2024timecma} separately encodes each modality and then employs a cross-attention alignment. Despite these advances, LLMs remain pretrained on predominantly textual corpora, leading to sparse numeric representations and limited numerical reasoning ability~\cite{huang2025survey, spathis2024first}. The imbalance in pre-training data raises concerns about the models’ ability to capture the fine-grained quantitative structures essential for reliable time series forecasting.

\subsection{Prompt Learning}
Prompt learning replaces fixed prompts with small set of trainable parameters to adapt LLMs to downstream tasks. For example, Prompt Tuning appends learnable soft prompts to input embeddings of a frozen LLM~\cite{lester2021power}, while Prefix Tuning inserts prefix vectors at each Transformer layer~\cite{ding2023parameter}, expanding capacity without altering model weights. In time series, TEMPO \cite{cao2023tempo} integrates seasonal and trend decomposition with semi-soft prompt to improve forecasting. LSTPrompt \cite{liu2024lstprompt} unlocks the forecasting potential of LLMs by designing task-specific prompts tailored to various time series forecasting tasks. Other work has shown that combining chain-of-thought prompting with minimal fine-tuning enables GPT-4 to detect anomalies in time series data~\cite{dong2024can}. Overall, prompt learning offers a lightweight and effective mechanism to leverage LLM knowledge for time series tasks.

\subsection{Multimodal Alignment}
Multimodal alignment has been widely studied, particularly in vision-language models, where frameworks such as CLIP~\cite{radford2021learning} and ALIGN~\cite{jia2021scaling} employ contrastive learning to project images and texts into a shared embedding space. In time series, the continuous, high-frequency, and noisy nature of data poses challenges for alignment with discrete, semantically rich text. To support research in this area, benchmarks such as the Time‐MMD dataset~\cite{liu2024time} provide large‐scale multimodal time‐series corpora across diverse domains. Architecturally, methods like Time‐CMA~\cite{liu2024timecma} leverage cross‐attention to fuse modalities, TEST~\cite{sun2023test} employs two complementary contrastive strategies to align time series embeddings with text prototypes, and CALF~\cite{liu2025calf} introduces multiple cross‐modal matching strategies for alignment. Despite these advances, the modality gap between continuous signals and rich linguistic representations makes existing methods computationally heavy and difficult to optimize.

\begin{figure*}[ht]
  \centering
  \includegraphics[width=\linewidth, trim = 0.5cm 1cm 1cm 4cm, clip]{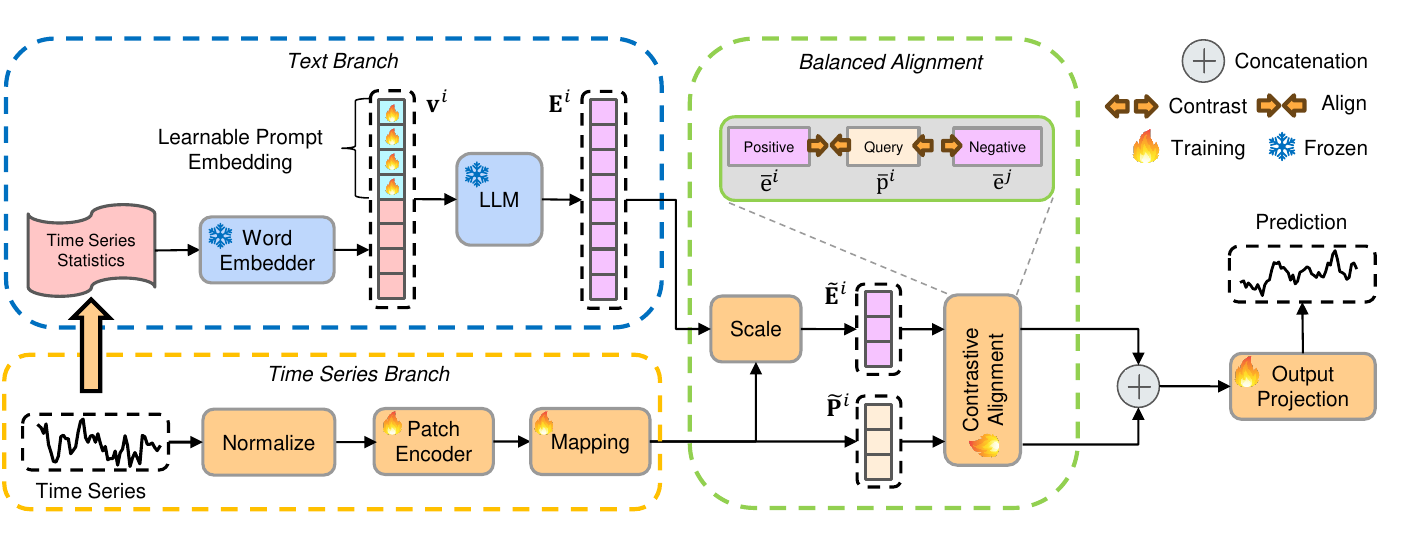}
  \vspace{-3.2em}
  \caption{Overview of BALM-TSF. The model has two branches: (1) the Text Branch constructs a prompt embedding by combining learnable prompt embedding with word-embedded time series statistics, then encodes it with a frozen LLM; (2) the Time Series Branch normalizes raw data, applies a patch encoder, and projects features into the LLM’s hidden space. In the Balanced Alignment module, textual embeddings are rescaled and then aligned with time series embeddings via a contrastive learning. The aligned two representations are concatenated and fed to a lightweight projection head for final prediction.}
  \label{fig1}
\end{figure*}

\section{Method}
\paragraph{Overview} 
As illustrated in Figure~\ref{fig1}, BALM-TSF adopts a dual-branch architecture with an alignment module to handle the heterogeneous modalities of time series and text. The model has two branches: the \textbf{Time Series Branch} normalizes the input and applies a patch encoder to project it into a compact embedding. The \textbf{Text Branch} encodes statistical information of the input time series in the form of prompts, which are combined with learnable prompts and fed into a frozen GPT-2 to inject domain-specific semantics into the textual embedding space. Subsequently, BALM-TSF scales and aligns the output of the text branch with the time series embedding, which are then concatenated to produce the final prediction. Compared to prior work, our design (1) captures temporal patterns from a language perspective, (2) alleviates modality imbalance between continuous series and discrete text, and (3) avoids excessive parameters and complex alignment. As a result, it achieves more efficient and accurate forecasting. Details of each component are presented in the following sections.

\begin{figure}[h]
  \centering
  \includegraphics[width=0.8\linewidth, trim = 6cm 5cm 6cm 5cm, clip]{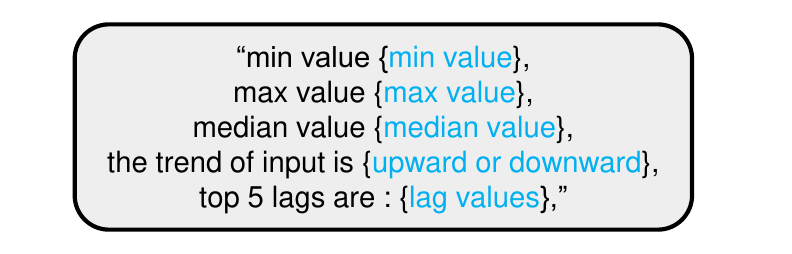}
  \vspace{-1.3em}
  \caption{Statistical prompt template for time series.}
  \label{fig_prompt}
\end{figure}

\subsection{Problem Formulation}
 We first define the multivariate time series forecasting problem. Previous timestamps of multivariate time series at time step $t$ are denoted as $ \mathbf{X}_{t-L+1:t} = \{X_{t-L+1}, X_{t-L+2}, ..., X_t\} \in \mathbb{R}^{L\times N}$, where $L$ is the length of the look-back window of the $N$ variables. The goal of multivariate time series forecasting is to predict the values for the next H 
 timestamps $\hat{\mathbf{X}}_{t+1:t+H} = \{\hat{X}_{t+1}, \hat{X}_{t+2}, ..., \hat{X}_{t+H}\} $. In addition, the prompt, a time series data corresponding instruction in text form is denoted by $\mathbf{V}$. The LLM-based forecaster $F$ with parameters $\Phi$ is trained by previous $L$-length time series and prompt $\mathbf{V}$ for the future H timestamps. The task is denoted as:
\begin{equation}
\hat{\mathbf{X}}_{t+1:t+H} = F(\mathbf{X}_{t-L+1:t};\mathbf{V};\Phi)
\end{equation}

\subsection{Time Series Branch}
 This branch is for encoding the time series itself, ensuring that domain-specific patterns can be learned.
\subsubsection{Normalize} To mitigate distribution shifts in time series inputs, each channel is first standardized to zero mean and unit variance via reversible instance normalization (RevIN) \cite{kim2021reversible}.
\subsubsection{Patch Encoder} We follow PatchTST \cite{Yuqietal-2023-PatchTST} to extract local semantics from each normalized variable of the time series. We first partition the time series variable into overlapping (or non-overlapping) 
patches of patch length \(L_P\) with stride \(S\), then embed them to get a patch representation. For the $i$-th channel time series variable \(X^{(i)}\) (equal to a variate), we obtain 
$\mathbf{P}^i \in\mathbb{R}^{N_P\times d_m}$ 
, where $d_m$ is patching embedding size and the number of patches $N_P$ is
\begin{equation}
N_P = \Big\lfloor \frac{L - L_P}{S} \Big\rfloor + 2.
\label{patchnums}
\end{equation}

\subsubsection{Mapping.} To enable later interaction with textual embeddings, we project the patch embeddings into the hidden dimension of the LLM using a linear mapping with trainable weights $W_m \in \mathbb{R}^{d_m \times d_{\mathrm{LLM}}}$ and bias $b_m \in \mathbb{R}^{d_{\mathrm{LLM}}}$:
\begin{equation}
\tilde{\mathbf{P}}^i = \mathbf{P}^i W_m + b_m \in \mathbb{R}^{N_P \times d_{\mathrm{LLM}}}
\end{equation}
where \(d_{\mathrm{LLM}}\) is the hidden size of the LLM output tokens.

\subsection{Text Branch}
In this branch, a frozen pre-trained LLM is used to extract informative semantic representations from time-series–related prompts. To adapt the LLM to our forecasting task, we design a soft prompt embedding $\mathbf{v}^i$, which incorporates both statistical and learnable prompt embeddings. The resulting prompt is then encoded by the LLM to generate textual embeddings.

\subsubsection{Prompt Design} We formulate our prompt as a combination of two components: statistical prompt and learnable prompt. 
\paragraph{Statistical Prompt} Following the statistical prompt design of Time-LLM~\cite{jin2023time}, we extract descriptive features from each time series, including the minimum, maximum, and median values, the overall trend, and the lag features, which are derived by computing the autocorrelation with fast Fourier transform (FFT) and selecting the five lag positions with the strongest correlations. The overall statistical prompt design is illustrated in Figure~\ref{fig_prompt}. The prompts follow a fixed template, with numeric values updated to reflect the current time series. This design avoids the volatility of feeding raw sequences into the LLM~\cite{liu2024timecma}, while still introducing dynamic, semantically rich content that adapts to the data and mitigates semantic imbalance. Moreover, standardizing the prompt structure with a fixed set of summary statistics allows downstream alignment to more reliably distinguish time series with similar summaries. The statistical prompt $\mathbf{V}^i_{\mathrm{stat}}$, derived from the $i$-th variable, is first tokenized and then passed through the LLM’s frozen word‐embedding layer to produce its input representation $\mathbf{v}^i_{\mathrm{stat}}$. We abbreviate this two‐step process as:
\begin{equation}
\mathbf{v^i}_{\mathrm{stat}} = \mathrm{WordEmbedder}(\mathbf{V^i}_{\mathrm{stat}}),
\end{equation}
where $\mathbf{V^i}_{\mathrm{stat}}$ denotes the textual prompt of the statistical information, and $\mathbf{v^i}_{\mathrm{stat}}$ is the corresponding embedded vector.

\paragraph{Learnable Prompt} The learnable prompt embedding $\mathbf{v}^i_{\mathrm{learn}}$ is randomly initialized without any prior distributional assumption. During training, $\mathbf{v}^i_{\mathrm{learn}}$ is updated via back-propagation to capture task-specific patterns. It is concatenated with the statistical embedding to form the final prompt embedding:
\begin{equation}
\mathbf{v}^i = [\mathbf{v}^i_{\mathrm{learn}};  \mathbf{v}^i_{\mathrm{stat}}].
\end{equation}

\subsubsection{LLM Encoder}
Pre-trained LLMs are trained on large-scale textual corpora and are capable of extracting rich semantic representations from prompts. In our framework, we adopt GPT-2 as the LLM backbone due to its lightweight architecture and strong modeling capacity. Specifically, GPT-2 encodes the input sequence through a stack of masked self-attention and feed-forward layers. Given the prompt embedding $\mathbf{v}^i$ as input, the LLM generates the textual embedding $\mathbf{E}^i $ as follows:
\begin{equation}
\mathbf{E}^i = \mathrm{LLM}(\mathbf{v}^i) \in \mathbb{R}^{N_T \times d_{\mathrm{LLM}}},
\end{equation}
where $N_T$ denotes the token length of $\mathbf{v}^i$, and $d_{\mathrm{LLM}}$ is the hidden size of LLM.

\subsection{Balanced Multimodal Alignment}
To address the distributional imbalance and semantics imbalance between textual and time series embeddings, we propose a balanced multimodal alignment strategy that combines value-level scaling with semantic-level contrastive alignment.
\subsubsection{Scale} To better facilitate alignment between the textual embeddings and time series embeddings, we apply a two-step scaling strategy based on the prediction horizon. Recent work shows that tokens differ in their impact on language model training~\cite{lin2024not, haviv2022transformer}; due to masked self-attention, final tokens in a prompt capture richer context than earlier ones. Since the primary role of the textual modality in time series forecasting is to provide auxiliary semantic information, it is important to avoid excessive textual influence, especially in short-horizon prediction where the temporal patterns dominate. Conversely, for long-horizon forecasting, more textual information is beneficial to supplement the weakened temporal signals. Thus, we truncate textual embeddings to adaptively retain the most relevant final tokens based on the prediction horizon.

Specifically, given the number of patches $N_P$, prediction length $H$, and total sequence length $L$, we adaptively determine the number of retained tokens $N_E$ as follows:
\begin{equation}
N_E = \min\left(N_P,\;\left\lfloor \frac{N_P \cdot H}{L} \right\rfloor\right),
\label{trunc}
\end{equation}
and truncate the embedding sequence accordingly:
\begin{equation}
\mathbf{E}_{\mathrm{trunc}}^i = \mathbf{E}^i_{[N_T-N_E:N_T]} \in \mathbb{R}^{N_E \times d_{\mathrm{LLM}}},
\end{equation}
where $\mathbf{E}_{\mathrm{trunc}}^i$ denotes the truncated textual embedding obtained by retaining the last $N_E$ tokens of $\mathbf{E}^i$. This strategy adaptively truncates the prompt text based on the prediction horizon, allowing more semantic information to be incorporated as the forecast length increases. At the same time, it ensures that the retained textual information never exceeds that of the time series itself. Thus, it maintains a balanced integration of both modalities. 

Next, we rescale the textual embeddings to match the numerical range of the time series modality. As noted earlier, the distributions of these embeddings differ substantially: time series representations are typically normalized, whereas textual embeddings retain higher variance. To address this, we compute a scaling factor:
\begin{equation}
\alpha = \frac{\mathrm{STD}_{\mathrm{time}}}{\mathrm{STD}_{\mathrm{text}}},
\end{equation}
where \(\mathrm{STD}_{\mathrm{text}}\) and \(\mathrm{STD}_{\mathrm{time}}\) are the standard deviations of the truncated textual and time series embeddings, respectively (computed along the token dimension). We then apply:
\begin{equation}
\tilde{\mathbf{E}}^i = \alpha \,\mathbf{E}_{\mathrm{trunc}}^i \in \mathbb{R}^{N_E \times d_{\mathrm{LLM}}}.
\end{equation}
This step enforces value‐level distributional consistency between modalities, thereby facilitating the subsequent alignment.

\subsubsection{Semantic Alignment} To encourage effective semantic alignment between the scaled textual embedding $\tilde{\mathbf{E}}^i$ with its temporal counterpart $\tilde{\mathbf{P}}^i$, we adopt a contrastive learning objective based on InfoNCE loss \cite{oord2018representation}. We average over the token dimension of $\tilde{\mathbf{E}}^i$ and $\tilde{\mathbf{P}}^i$ to obtain their global representations:
\begin{equation}
\mathbf{e}^i = \frac{1}{N_E} \sum_{m=1}^{N_E} \tilde{\mathbf{E}}^i_m, \quad
\mathbf{p}^i = \frac{1}{N_P} \sum_{n=1}^{N_P} \tilde{\mathbf{P}}^i_n,
\end{equation}
where $m$ denotes $m$-th token in $\tilde{\mathbf{E}}^i$ and $n$ denotes $n$-th token in $\tilde{\mathbf{P}}^i$. Then we normalize them by $\ell_2$ normalization:
\begin{equation}
\bar{\mathbf{e}}^i = \frac{\mathbf{e}^i}{\|\mathbf{e}^i\|}, \quad
\bar{\mathbf{p}}^i = \frac{\mathbf{p}^i}{\|\mathbf{p}^i\|}.
\end{equation}
$\bar{\mathbf{p}}^i$ served as a query in contrastive learning, as shown in Figure \ref{fig1}. The positive sample $\bar{\mathbf{e}}^i$ is corresponding statistical information embedding of $\bar{\mathbf{p}}^i$. Thus, each pair $(\bar{\mathbf{p}}^i, \bar{\mathbf{e}}^i)$ is treated as a positive pair, while $(\bar{\mathbf{p}}^i, \bar{\mathbf{e}}^j)$ for $j \ne i$ serve as negative pairs. A learnable temperature $\tau = \exp(\theta)$ is introduced to scale similarity. Given K samples, the final alignment loss is computed as:
\begin{equation}
\mathcal{L}_{align} = -\frac{1}{K} \sum_{i=1}^{K} \log \frac{\exp(\bar{\mathbf{p}}^i \cdot \bar{\mathbf{e}}^i / \tau)}{\sum_{j=1}^{K} \exp(\bar{\mathbf{p}}^i \cdot \bar{\mathbf{e}}^j / \tau)}.
\end{equation}
Leveraging the preceding scaling step, this semantic alignment promotes each time series embedding to be maximally similar to its textual embedding with corresponding statistical information while remaining dissimilar to other samples.

\subsection{Objective Function}
To leverage both modalities for forecasting, we concatenate the textual and time series embeddings into a joint representation $\mathbf{Z}_{\text{out}} = [\tilde{\mathbf{E}}^i; \tilde{\mathbf{P}}^i ]$, which is then flattened and projected via a linear layer to obtain the prediction. We use Mean Squared Error (MSE) to calculate forecasting task loss $\mathcal{L}_{\text{task}}$ with ground truth.
\paragraph{Final Objective} The overall objective of BALM-TSF combines the forecasting task loss and alignment loss:
\begin{equation}
\mathcal{L} = \mathcal{L}_{\text{task}} + \lambda \mathcal{L}_{\text{align}},
\end{equation}
 The hyperparameter $\lambda$ controls the trade-off between forecasting loss and alignment loss.

\section{Experiments}
\begin{table*}[ht]
    \centering
    \caption{Long-term forecasting results of BALM-TSF and baseline methods. A look-back window of 512 is used with prediction horizons $H\in\{96,192,336,720\}$. Performance is measured by MSE and MAE, with best values in bold and second-best underlined.}

    \label{tab:LT}
    \small
    \setlength{\tabcolsep}{1.2mm}
    \begin{tabular}{c|c|cc|cc|cc|cc|cc|cc|cc|cc|cc}
        \toprule
        \multicolumn{2}{c|}{Model} & \multicolumn{2}{c|}{BALM-TSF} & \multicolumn{2}{c|}{Time-LLM} & \multicolumn{2}{c|}{TimeCMA} & \multicolumn{2}{c|}{GPT4TS} & \multicolumn{2}{c|}{UniTime} & \multicolumn{2}{c|}{iTransformer} & \multicolumn{2}{c|}{PatchTST} & \multicolumn{2}{c|}{DLinear} & \multicolumn{2}{c}{TimesNet} \\
        \cmidrule(lr){1-20}
        \multicolumn{2}{c|}{Metric} & MSE & MAE & MSE & MAE & MSE & MAE & MSE & MAE & MSE & MAE & MSE & MAE & MSE & MAE & MSE & MAE & MSE & MAE \\
        \midrule
        \multirow{5}{*}{\rotatebox{90}{ETTh1}} 
        & 96  &\textbf{0.365} &\textbf{0.392} &0.387 &0.411 &0.409 &0.431 &0.381 &0.404 &0.487 &0.478 &0.399 &0.423 &0.390 &0.416 &\underline{0.367} &\underline{0.396} &0.448 &0.460  \\
        & 192 &\textbf{0.398} &\textbf{0.412} &0.419 &0.434 &0.435 &0.444 &0.415 &0.424 &0.501 &0.488 &0.426 &0.442 &0.454 &0.460 &\underline{0.402} &\underline{0.419} &0.495 &0.491  \\
        & 336 &\textbf{0.423} &\textbf{0.428} &0.451 &0.458 &0.435 &0.450 &0.433 &\underline{0.436} &0.505 &0.497 &0.463 &0.466 &0.483 &0.479 &\underline{0.429} &0.441 &0.536 &0.517  \\
        & 720 &\textbf{0.440} &\textbf{0.464} &\underline{0.443} &0.465 &0.450 &0.472 &0.455 &\underline{0.467} &0.549 &0.529 &0.679 &0.598 &0.672 &0.577 &0.475 &0.497 &0.578 &0.539  \\
        \cmidrule(){2-20}
        & Avg &\textbf{0.406} &\textbf{0.424} &0.425 &0.442 &0.432 &0.449 &0.421 &\underline{0.433} &0.510 &0.498 &0.492 &0.482 &0.500 &0.483 &\underline{0.418} &0.438 &0.514 &0.502  \\
        \midrule
        \multirow{5}{*}{\rotatebox{90}{ETTh2}} 
        & 96 &\textbf{0.267} &\textbf{0.335} &0.303 &0.357 &0.379 &0.412 &\underline{0.292} &\underline{0.353} &0.403 &0.418 &0.302 &0.358 &0.313 &0.370 &0.300 &0.365 &0.361 &0.408  \\
        & 192 &\textbf{0.328} &\textbf{0.373} &\underline{0.359} &\underline{0.391} &0.374 &0.415 &0.366 &0.394 &0.510 &0.448 &0.373 &0.404 &0.391 &0.417 &0.396 &0.427 &0.409 &0.437  \\
        & 336 &\textbf{0.349} &\textbf{0.397} &\underline{0.375} &\underline{0.412} &0.447 &0.463 &0.385 &0.417 &0.742 &0.471 &0.418 &0.435 &0.446 &0.455 &0.498 &0.491 &0.394 &0.438  \\
        & 720 &\textbf{0.390} &\textbf{0.432} &\underline{0.415} &\underline{0.446} &0.497 &0.496 &0.434 &0.456 &1.335 &0.523 &0.447 &0.466 &0.501 &0.497 &0.806 &0.636 &0.485 &0.490  \\
        \cmidrule(){2-20}
        & Avg &\textbf{0.333} &\textbf{0.384} &\underline{0.363} &\underline{0.402} &0.424 &0.446 &0.369 &0.405 &0.747 &0.465 &0.385 &0.416 &0.413 &0.435 &0.500 &0.480 &0.412 &0.443  \\
        \midrule
        \multirow{5}{*}{\rotatebox{90}{ETTm1}} 
        & 96 &\underline{0.304} &\textbf{0.347} &0.337 &0.372 &0.319 &0.374 &\textbf{0.295} &0.352 &0.380 &0.407 &0.310 &0.364 &0.304 &0.354 &0.305 &\underline{0.349} &0.349 &0.381  \\
        & 192 &\textbf{0.336} &\textbf{0.367} &0.360 &0.386 &0.356 &0.395 &\underline{0.337} &0.375 &0.409 &0.421 &0.348 &0.387 &0.359 &0.394 &0.337 &\underline{0.368} &0.480 &0.449  \\
        & 336 &\textbf{0.365} &\textbf{0.384} &0.388 &0.401 &0.383 &0.410 &0.369 &0.394 &0.430 &0.433 &0.378 &0.404 &0.379 &0.405 &\underline{0.366} &\underline{0.385} &0.410 &0.426  \\
        & 720 &\underline{0.420} &\textbf{0.417} &0.433 &0.427 &0.424 &0.433 &\textbf{0.418} &0.422 &0.488 &0.466 &0.440 &0.441 &0.444 &0.448 &0.420 &\underline{0.417} &0.468 &0.462  \\
        \cmidrule(){2-20}
        & Avg &\underline{0.356} &\textbf{0.379} &0.379 &0.397 &0.371 &0.403 &\textbf{0.354} &0.386 &0.427 &0.432 &0.369 &0.399 &0.371 &0.400 &0.357 &\underline{0.380} &0.427 &0.430  \\
        \midrule
        \multirow{5}{*}{\rotatebox{90}{ETTm2}} 
        & 96 &\textbf{0.165} &\textbf{0.255} &0.181 &0.274 &0.191 &0.277 &0.171 &0.266 &0.249 &0.323 &0.180 &0.272 &0.182 &0.272 &\underline{0.166} &\underline{0.261} &0.194 &0.281  \\
        & 192 &\textbf{0.218} &\textbf{0.292} &0.235 &0.309 &0.258 &0.327 &0.228 &0.304 &0.299 &0.353 &0.242 &0.313 &0.236 &0.309 &\underline{0.224} &\underline{0.303} &0.260 &0.325  \\
        & 336 &\textbf{0.270} &\textbf{0.326} &\underline{0.283} &0.340 &0.317 &0.360 &0.283 &\underline{0.339} &0.349 &0.383 &0.294 &0.347 &0.293 &0.346 &0.296 &0.359 &0.321 &0.368  \\
        & 720 &\textbf{0.366} &\textbf{0.388} &\underline{0.368} &\underline{0.390} &0.397 &0.411 &0.378 &0.399 &0.436 &0.434 &0.373 &0.395 &0.396 &0.408 &0.410 &0.432 &0.422 &0.421  \\
        \cmidrule(){2-20}
        & Avg &\textbf{0.255} &\textbf{0.316} &0.267 &0.328 &0.290 &0.343 &\underline{0.265} &\underline{0.327} &0.333 &0.373 &0.272 &0.332 &0.277 &0.334 &0.274 &0.339 &0.299 &0.349  \\
        \midrule
        \multirow{5}{*}{\rotatebox{90}{Weather}} 
        & 96 &0.172 &0.226 &\underline{0.157} &\underline{0.209} &0.165 &0.220 &0.149 &\textbf{0.200} &0.195 &0.293 &0.169 &0.220 &\textbf{0.155} &0.211 &0.171 &0.232 &0.164 &0.220  \\
        & 192 &0.213 &0.260 &\underline{0.198} &\underline{0.247} &0.201 &0.253 &\textbf{0.194} &\textbf{0.243} &0.240 &0.273 &0.212 &0.257 &0.208 &0.259 &0.215 &0.274 &0.221 &0.270  \\
        & 336 &0.262 &0.297 &0.250 &0.287 &\textbf{0.247} &0.288 &\underline{0.248} &\textbf{0.285} &0.292 &0.305 &0.267 &0.295 &0.248 &\underline{0.286} &0.259 &0.292 &0.281 &0.313  \\
        & 720 &0.323 &0.339 &\underline{0.317} &\underline{0.334} &\textbf{0.311} &\textbf{0.334} &0.320 &0.336 &0.357 &0.345 &0.334 &0.341 &0.318 &0.336 &0.320 &0.359 &0.344 &0.356  \\
        \cmidrule(){2-20}
        & Avg &0.243 &0.281 &\underline{0.230} &\underline{0.269} &0.231 &0.274 &\textbf{0.228} &\textbf{0.266} &0.271 &0.304 &0.246 &0.278 &0.232 &0.273 &0.241 &0.289 &0.253 &0.290  \\
        \midrule
        \multirow{5}{*}{\rotatebox{90}{Exchange}} 
        & 96 &\textbf{0.091} &\textbf{0.212} &0.107 &0.235 &0.250 &0.377 &\underline{0.097} &\underline{0.218} &0.199 &0.329 &0.107 &0.233 &0.216 &0.246 &0.117 &0.259 &0.247 &0.369  \\
        & 192 &\textbf{0.187} &\textbf{0.307} &0.249 &0.362 &0.628 &0.605 &0.218 &0.333 &0.311 &0.416 &\underline{0.208} &\underline{0.327} &0.209 &1.274 &0.234 &0.368 &0.375 &0.456  \\
        & 336 &\textbf{0.348} &\textbf{0.429} &0.465 &0.502 &0.799 &0.677 &0.399 &\underline{0.459} &0.495 &0.529 &0.419 &0.480 &\underline{0.378} &0.453 &0.627 &0.620 &0.622 &0.601  \\
        & 720 &\textbf{0.896} &\textbf{0.710} &1.192 &0.823 &2.427 &1.210 &1.106 &0.779 &1.129 &0.801 &\underline{1.020} &\underline{0.765} &1.274 &0.834 &1.170 &0.827 &1.515 &0.928  \\
        \cmidrule(){2-20}
        & Avg &\textbf{0.380} &\textbf{0.415} &0.503 &0.481 &1.026 &0.717 &0.455 &\underline{0.447} &0.534 &0.519 &\underline{0.439} &0.451 &0.519 &0.702 &0.537 &0.519 &0.690 &0.588  \\
        \bottomrule
    \end{tabular}
\end{table*} 

\begin{table*}[h]
    \centering
    \caption{Few-shot forecasting results on 10\% of the training data. A look-back window of 512 is used with prediction horizons $H\in\{96,192,336,720\}$. Performance is measured by MSE and MAE, with best values in bold and second-best underlined.}

    \label{tab:FT}
    \small
    \setlength{\tabcolsep}{1.2mm}
    \begin{tabular}{c|c|cc|cc|cc|cc|cc|cc|cc|cc|cc}
        \toprule
        \multicolumn{2}{c|}{Model} & \multicolumn{2}{c|}{BALM-TSF} & \multicolumn{2}{c|}{Time-LLM} & \multicolumn{2}{c|}{TimeCMA} & \multicolumn{2}{c|}{GPT4TS} & \multicolumn{2}{c|}{UniTime} & \multicolumn{2}{c|}{iTransformer} & \multicolumn{2}{c|}{PatchTST} & \multicolumn{2}{c|}{DLinear} & \multicolumn{2}{c}{TimesNet} \\
        \cmidrule(lr){1-20}
        \multicolumn{2}{c|}{Metric} & MSE & MAE & MSE & MAE & MSE & MAE & MSE & MAE & MSE & MAE & MSE & MAE & MSE & MAE & MSE & MAE & MSE & MAE \\
        \midrule
        \multirow{5}{*}{\rotatebox{90}{ETTh1}} 
        & 96  &\underline{0.479}  &\underline{0.464}  &0.541  &0.495  &0.701  &0.560  &0.588  &0.494  &0.751  &0.577  &0.589  &0.520  &0.546  &0.506  &\textbf{0.450}  &\textbf{0.452}  &0.869  &0.636  \\
        & 192 &\underline{0.518} &\underline{0.492} &0.690 &0.569 &0.761 &0.572 &0.715 &0.536 &0.770 &0.586 &0.661 &0.546 &0.629 &0.544 &\textbf{0.478} &\textbf{0.469} &0.883 &0.652 \\
        & 336 &0.818 &0.646 &1.034 &0.715 &0.754 &0.586 &0.759 &\underline{0.571} &0.798 &0.600 &\underline{0.731} &0.636 &0.724 &0.599 &\textbf{0.500} &\textbf{0.488} &0.866 &0.640 \\
        & 720 &0.918 &0.700 &0.957 &0.663 &0.753 &0.613 &\underline{0.745} &\underline{0.610} &0.904 &0.644 &0.849 &0.650 &0.798 &0.646 &\textbf{0.692} &\textbf{0.599} &0.883 &0.663 \\
        \cmidrule(){2-20}
        & Avg &0.683 &0.576 &0.805 &0.610 &0.742 &0.583 &0.702 &\underline{0.552} &0.806 &0.602 &0.708 &0.588 &\underline{0.674} &0.574 &\textbf{0.530} &\textbf{0.502} &0.875 &0.648 \\
        \midrule
        \multirow{5}{*}{\rotatebox{90}{ETTh2}} 
        & 96 &\underline{0.330} &0.389 &0.373 &0.412 &0.383 &0.426 &\textbf{0.317} &\textbf{0.377} &0.406 &0.432 &0.358 &0.401 &0.384 &0.425 &0.331 &\underline{0.384} &0.464 &0.476 \\
        & 192 &0.431 &0.444 &0.466 &0.480 &0.441 &0.462 &0.418 &\underline{0.441} &0.443 &0.454 &\underline{0.420} &\textbf{0.438} &0.460 &0.456 &\textbf{0.413} &0.441 &0.481 &0.489  \\
        & 336 &0.533 &0.501 &0.572 &0.520 &0.447 &0.463 &0.459 &0.468 &0.462 &0.469 &\textbf{0.434} &\textbf{0.454} &0.493 &0.485 &\underline{0.438} &\underline{0.459} &0.463 &0.474  \\
        & 720 &0.546 &0.519 &0.481 &\underline{0.478} &0.481 &0.487 &0.722 &0.616 &0.512 &0.503 &0.491 &0.495 &\textbf{0.463} &\textbf{0.473} &\underline{0.466} &0.496 &0.503 &0.502  \\
        \cmidrule(){2-20}
        & Avg &0.460 &0.463 &0.473 &0.473 &0.438 &\underline{0.459} &0.479 &0.475 &0.456 &0.464 &\underline{0.426} &0.447 &0.450 &0.460 &\textbf{0.412} &\textbf{0.445} &0.478 &0.485  \\
        \midrule
        \multirow{5}{*}{\rotatebox{90}{ETTm1}} 
        & 96 &\textbf{0.359} &\textbf{0.385} &0.418 &0.418 &0.527 &0.484 &0.391 &0.404 &0.376 &0.403 &0.447 &0.447 &0.501 &0.463 &\underline{0.362} &\underline{0.396} &0.717 &0.573  \\
        & 192 &\textbf{0.386} &\textbf{0.401} &0.486 &0.451 &0.660 &0.551 &0.430 &0.423 &0.404 &0.418 &0.515 &0.486 &0.547 &0.478 &\underline{0.387} &\underline{0.411} &0.783 &0.605  \\
        & 336 &\underline{0.421} &\textbf{0.421} &0.520 &0.469 &0.768 &0.592 &0.467 &0.440 &0.427 &0.432 &0.584 &0.519 &0.579 &0.496 &\textbf{0.418} &\underline{0.430} &0.835 &0.599  \\
        & 720 &0.532 &0.488 &0.675 &0.550 &0.849 &0.633 &0.603 &0.512 &\textbf{0.484} &\textbf{0.463} &0.660 &0.558 &0.957 &0.643 &\underline{0.490} &\underline{0.476} &0.867 &0.619  \\
        \cmidrule(){2-20}
        & Avg &0.424 &\textbf{0.424} &0.525 &0.472 &0.701 &0.565 &0.473 &0.445 &\textbf{0.423} &0.429 &0.552 &0.502 &0.646 &0.520 &\textbf{0.414} &\underline{0.428} &0.801 &0.599  \\
        \midrule
        \multirow{5}{*}{\rotatebox{90}{ETTm2}} 
        & 96 &\textbf{0.180} &\textbf{0.264} &0.222 &0.297 &0.254 &0.323 &0.191 &\underline{0.271} &0.248 &0.320 &0.203 &0.289 &0.218 &0.302 &\underline{0.207} &0.294 &0.282 &0.346  \\
        & 192 &\textbf{0.234} &\textbf{0.302} &0.267 &0.329 &0.293 &0.348 &\underline{0.251} &\underline{0.309} &0.298 &0.350 &0.258 &0.326 &0.283 &0.342 &0.265 &0.331 &0.323 &0.370  \\
        & 336 &\textbf{0.286} &\textbf{0.338} &0.328 &0.366 &0.347 &0.387 &0.311 &\underline{0.351} &0.347 &0.382 &\underline{0.309} &0.358 &0.342 &0.376 &0.318 &0.367 &0.382 &0.406  \\
        & 720 &0.446 &0.430 &0.438 &0.431 &0.483 &0.469 &\textbf{0.399} &\textbf{0.399} &0.433 &0.434 &0.411 &0.418 &0.468 &0.447 &\underline{0.400} &\underline{0.416} &0.500 &0.473  \\
        \cmidrule(){2-20}
        & Avg &\textbf{0.286} &\underline{0.333} &0.314 &0.356 &0.344 &0.382 &\underline{0.288} &\textbf{0.333} &0.331 &0.372 &0.295 &0.348 &0.328 &0.367 &0.298 &0.352 &0.372 &0.399  \\
        \bottomrule
    \end{tabular}
\end{table*}

\subsection{Experimental Setup}
\subsubsection{Datasets}
We train and evaluate the BALM-TSF model on six widely used multivariate time series datasets: ETTh1, ETTh2, ETTm1, ETTm2, Exchange, and Weather datasets \cite{zhou2021informer, lai2018modeling, wu2021autoformer}.

\subsubsection{Baselines}
We select eight the most advanced time series forecasting models for extensive evaluations. The four LLM-based forecasters are Time-LLM \cite{jin2023time} and UniTime \cite{liu2024unitime}, both of which feed textual embeddings and time series embeddings jointly into an LLM; TimeCMA \cite{liu2024timecma}, a dual-branch architecture that processes time series and text separately before alignment; and GPT4TS \cite{zhou2023one}, which leverages LLM to forecast directly from raw series data. The four non-LLM baselines include iTransformer \cite{liu2023itransformer}, PatchTST \cite{Yuqietal-2023-PatchTST}, DLinear \cite{Zeng2022AreTE}, TimesNet \cite{wu2023timesnet}, which together represent a diverse set of forecasting architectures, including Transformer-based, CNN-based, and MLP-based models

\subsubsection{Implementation Details}
We adhere to the experimental configurations as in Time-LLM \cite{jin2023time} across all baselines for fair comparison. We use GPT-2 \cite{radford2019language} with six Transformer layers as the LLM backbone for all LLM-based models. In BALM-TSF, stride is 8 and patch length is 16. We employ the Adam optimizer. The hyperparameter $\lambda$ is set to 1. All experiments are conducted on NVIDIA A100 GPUs and repeated three times, reporting averaged results.

\subsection{Long-term Forecasting}
\paragraph{Setup.} For long-term forecasting, We evaluated BALM-TSF across six datasets, which are ETTh1, ETTh2, ETTm1, ETTm2, Weather and Exchange. We fix the look-back window to 512 and evaluate the performance on four different horizons {96, 192, 336, 720}. The evaluation metrics include the mean square error (MSE) and the mean absolute error (MAE).

\paragraph{Results.}
Table~\ref{tab:LT} reports MSE and MAE for all methods across multiple datasets and forecasting horizons. BALM-TSF achieves the lowest or second-lowest error on all ETT and Exchange dataset–horizon combinations. Notably, compared to DLinear, which serves as the strongest non-LLM baseline, BALM-TSF achieves average reductions of \(\mathbf{11.1\%}\) in MSE and \(\mathbf{8.6\%}\) in MAE on all datasets. Furthermore, BALM-TSF also surpasses prior LLM-based forecasters, reducing the average MSE of Time-LLM, the strongest among existing LLM-based, reducing its MSE by \(\mathbf{8.9\%}\) and its MAE by \(\mathbf{5.8\%}\). These results confirm that our two-stage alignment, combining semantic contrastive learning with distributional scaling, enables BALM-TSF to extract richer, meaningful information from both text and time series. However, BALM-TSF sees smaller gains on the Weather dataset, one potential reason is our text branch uses only univariate statistics, while weather forecasting often requires richer external context. Future work will extend prompt design with auxiliary textual inputs to better capture high-frequency weather dynamics.

\subsection{Few-shot Forecasting}
\paragraph{Setup.} LLM-based models have shown superiority in few-shot learning tasks in the recent literature. We follow the Time-LLM \cite{jin2023time} setting to do few-shot forecasting with 10\% training data. We evaluate BALM-TSF across four datasets, which are ETTh1, ETTh2, ETTm1, ETTm2.

\paragraph{Results.}
Table~\ref{tab:FT} reports few-shot forecasting results for BALM-TSF and all baselines across four datasets. Overall, BALM-TSF consistently outperforms all prior LLM-based forecasters. In particular, it reduces the average MSE of the strongest LLM-based benchmark, GPT4TS, by $\mathbf{10.7}\%$ and its MAE by $\mathbf{5.8}\%$. Moreover, on the ETTm1 and ETTm2 datasets, BALM-TSF also surpasses the best non-LLM baseline, DLinear, achieving 6.3\% lower MSE and 4.6\% lower MAE. These findings underscore BALM-TSF’s advanced few-shot learning capability: by decoupling modalities into two branches and employing balanced alignment, it effectively harnesses the strong generalization of LLMs even with limited training samples.
\subsection{Ablation Study}
\begin{figure}[!htbp]
  \centering
  \includegraphics[width=\linewidth]{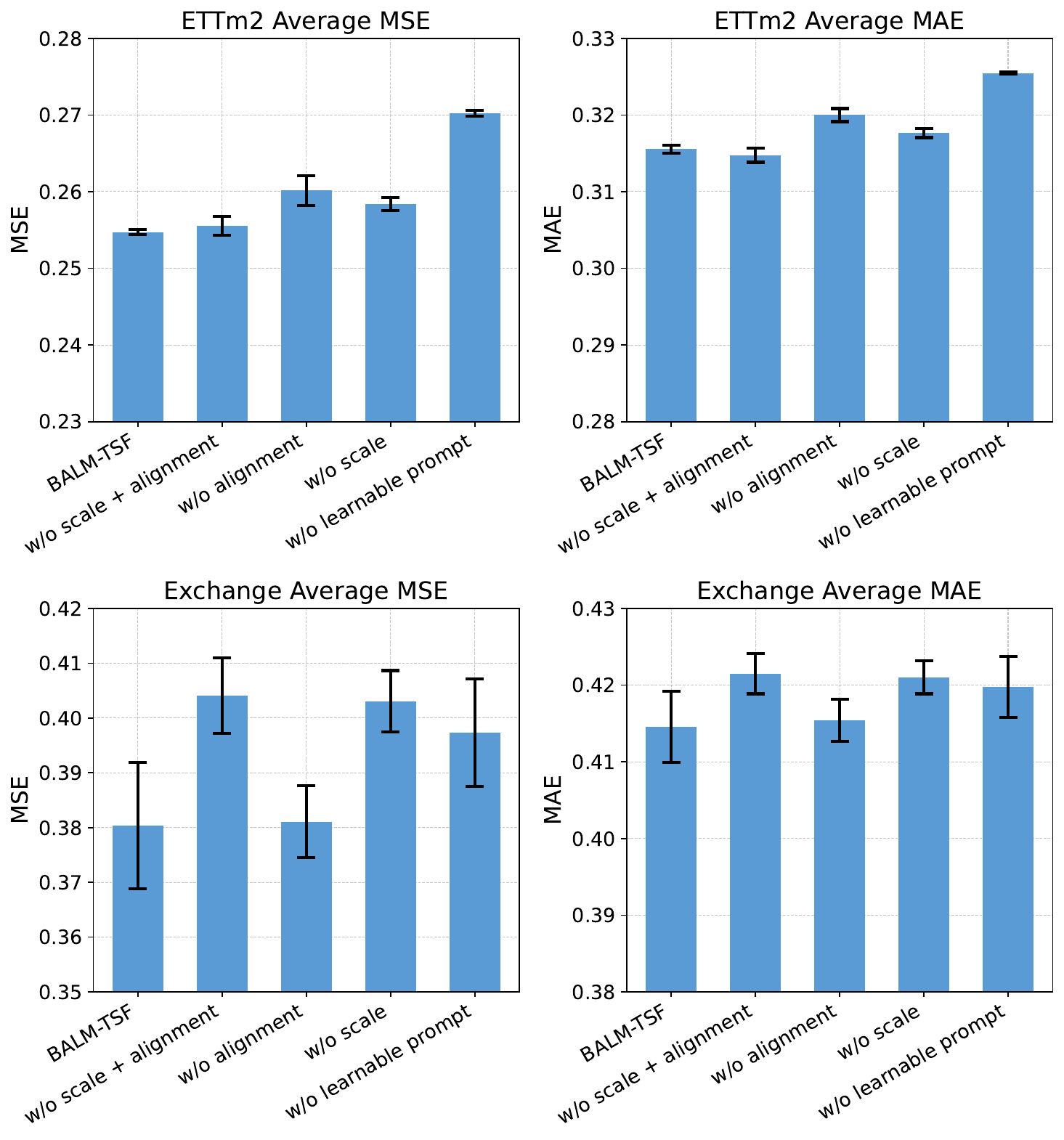}
  \vspace{-2em}
  \caption{Ablation study on the ETTm2 and Exchange datasets, reporting average MSE and MAE with error bars across all four forecasting horizons.}
  \label{fig:ablation}
\end{figure}

As shown in Figure~\ref{fig:ablation}, we assess the contribution of each BALM-TSF component by evaluating four ablated variants and reporting their average performance across all four horizons: (1) removing whole balanced alignment module ("w/o scale + alignment"), (2) omitting only contrastive alignment ( "w/o alignment"), (3) omitting only scaling ("w/o scale"), and (4) removing the learnable prompt ("w/o learnable prompt"). Each ablation results in a noticeable increase in both MSE and MAE on the ETTm2 and Exchange datasets, confirming the necessity of every component. The most substantial performance degradation is observed when the learnable prompt is removed on ETTm2, underscoring its importance in injecting semantic cues. 
Additionally, removing the entire balanced alignment module leads to particularly poor performance on the Exchange dataset, where the discrepancy between textual and time series embeddings still exists. Moreover, the error bars represent the standard deviation over three independent runs. On the ETTm2 dataset, we observe that removing the alignment and scaling modules leads to larger error bars compared to other variants, indicating that these components contribute to the overall stability of the model. These findings further demonstrate the effectiveness of the proposed two-stage balanced alignment mechanism and prompt design.

\begin{figure}[h]
  \centering
  \includegraphics[width=\linewidth]{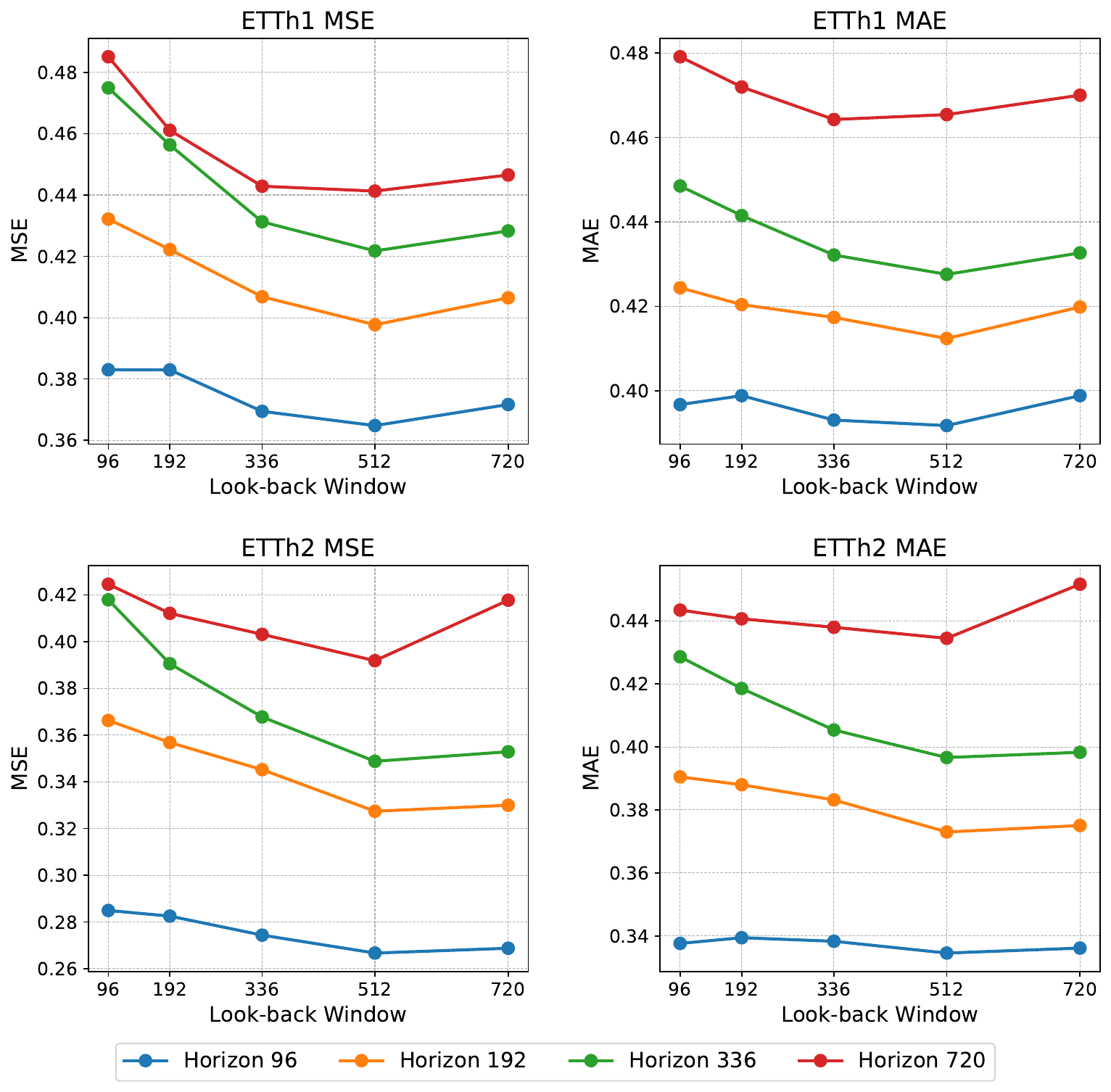}
  \vspace{-2em}
  \caption{Forecasting performance of BALM-TSF with varying look-back window $L\in\{96, 192, 336, 512, 720\}$.}
  \label{fig:seqlen}
\end{figure}
\begin{figure}[h]
  \centering
  \includegraphics[width=\linewidth]{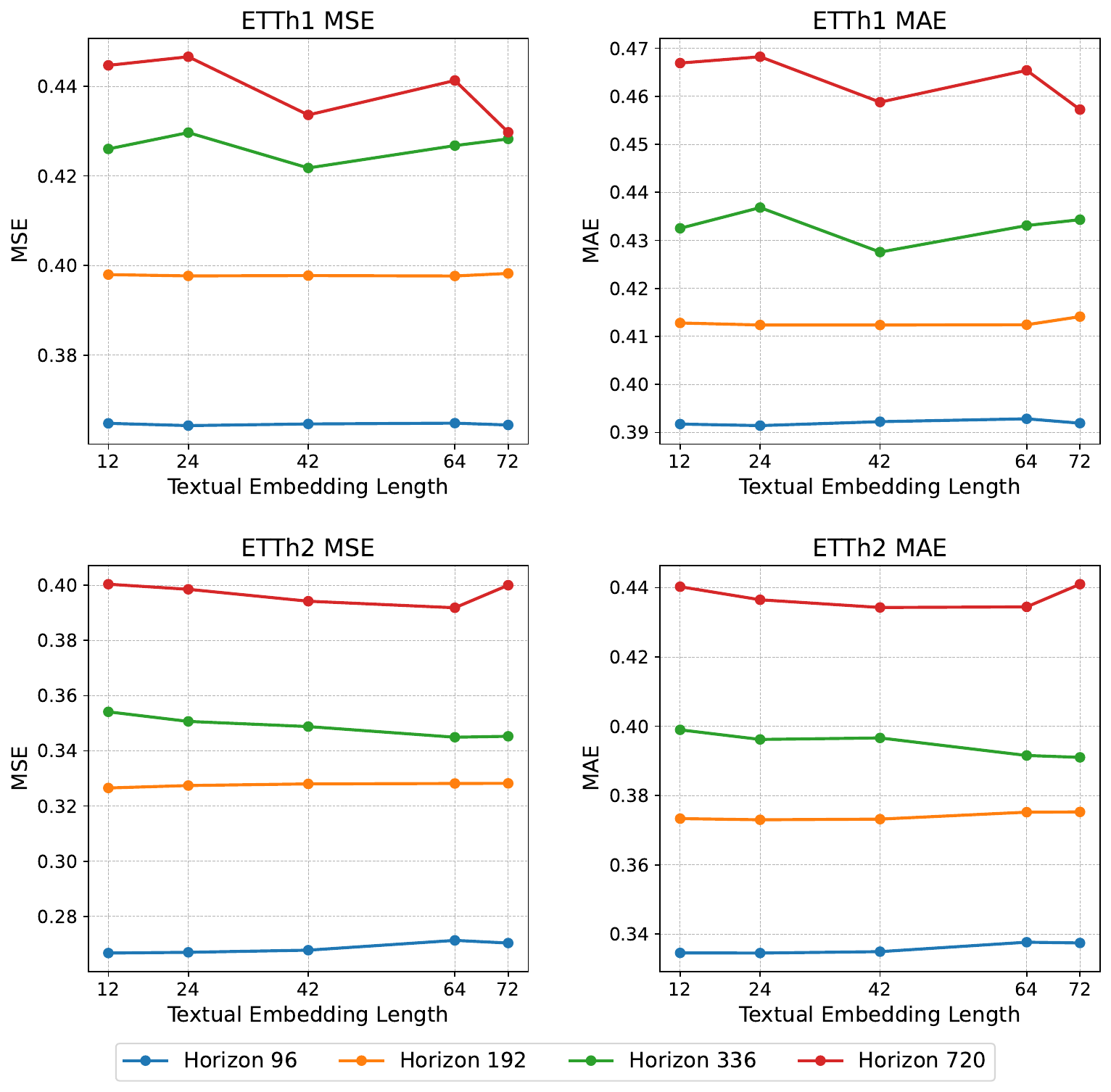}
  \vspace{-2em}
  \caption{Forecasting performance of BALM-TSF with varying retained textual embedding length $N_E\in\{12, 24, 42, 64, 72\}$.}
  \label{fig:textlen}
\end{figure}
\begin{figure}[h]
  \centering
  \includegraphics[width=1.01\linewidth]{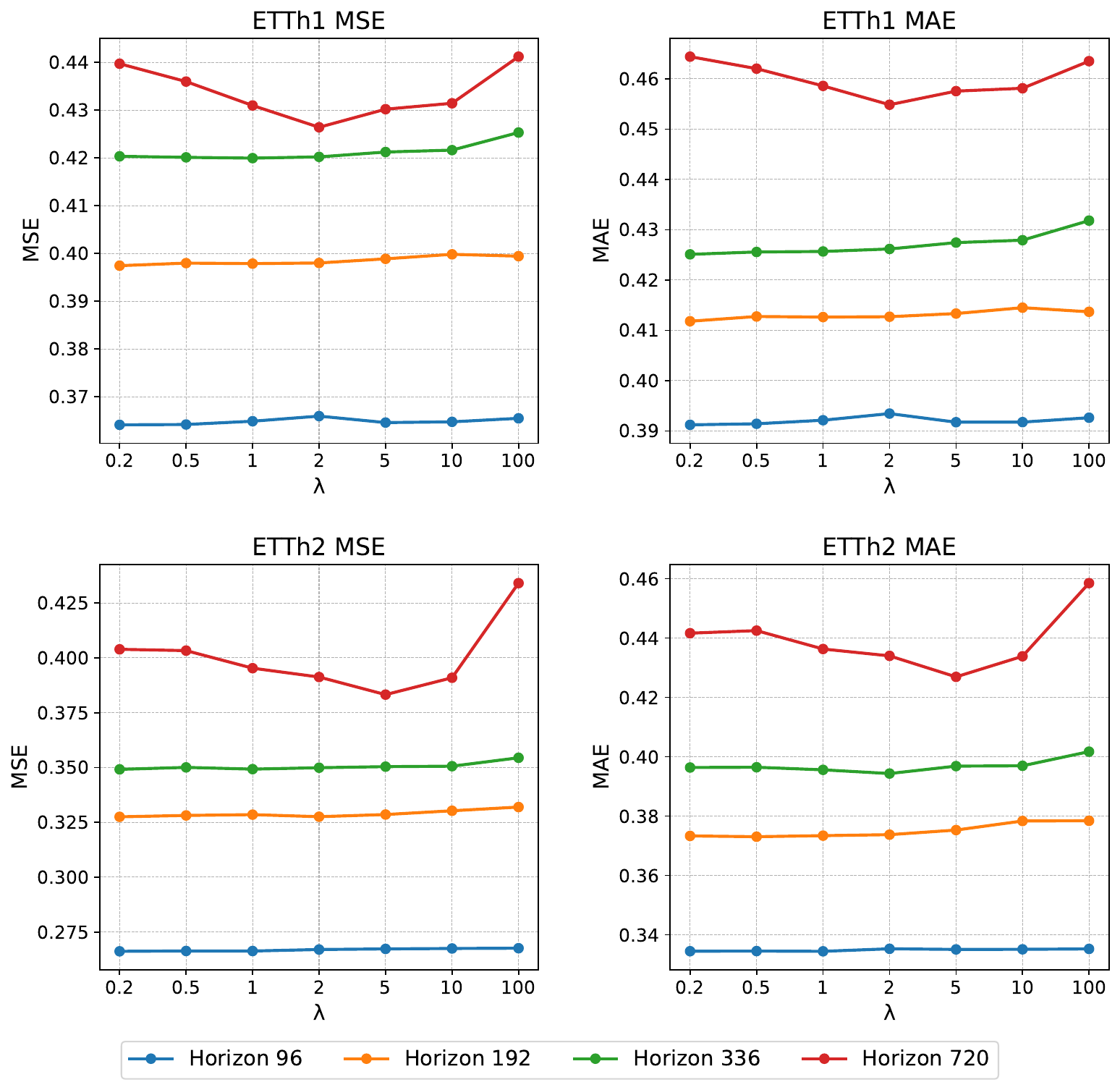}
  \vspace{-2em}
  \caption{Forecasting performance of BALM-TSF with varying $\lambda\in\{0.2, 0.5, 1, 2, 5, 10, 100\}$.}
  \label{fig:lamda}
\end{figure}

\subsection{Parameter Sensitivity Analysis}
\paragraph{Look-back Window}
 Figure \ref{fig:seqlen} reports the effect of different look-back window size $L\in\{96, 192, 336, 512, 720\}$ on ETTh datasets. Both datasets show a consistent pattern: performance improves as \(L\) increases up to \(L=512\), then degrades when extended to \(L=720\), with higher MSE/MAE at the longest window. This suggests that indiscriminately enlarging the look-back window may introduce redundant or weakly relevant context, making it harder for the model to capture the underlying complex temporal relationships.

\paragraph{Textual Embedding Length after Truncation}  
Figure~\ref{fig:textlen} reports the effect of varying Length of textual embeddings $N_E$ in the truncated textual embeddings on the forecasting performance of the ETTh datasets. Based on the patch count of 64 (Equation~\ref{patchnums}) and the truncation rule (Equation~\ref{trunc}), the retained textual embedding lengths corresponding to forecasting horizons of 96, 192, 336, and 720 steps are 12, 24, 42, and 64 tokens, respectively. We evaluate these values and additionally include 72 tokens, as $N_E \in \{12, 24, 42, 64, 72\}$. The results indicate that for short horizons (96 and 192), using more tokens degrades performance, whereas for long horizons (336 and 720), larger textual embeddings bring clear gains. This suggests that short-horizon forecasts rely mainly on intrinsic temporal signals, while long-horizon forecasts benefit from richer statistical context in the textual prompts. These findings highlight the horizon-dependent nature of the textual embedding length and support the adaptive truncation strategy in our scale module.

\paragraph{$\lambda$ in Loss Function}
Figure~\ref{fig:lamda} reports the impact of different $\lambda\in\{0.2, 0.5, 1, 2, 5, 10, 100\}$ on the forecasting performance of the ETTh datasets. As observed, forecasts with horizons of 96, 192, and 336 are relatively insensitive to increases in $\lambda$ within the range of 0.2 to 10, whereas the horizon of 512 benefits from increasing value and stabilizes afterwards. However, setting $\lambda$ to an excessively large value (e.g., 100) results in a clear performance drop across all horizons. These results indicate that different horizons require different degrees of alignment loss weighting. These observations offer some insights for future improvements and applications.

\begin{table*}
  \caption{Efficiency analysis of LLM-based models tested on the ETT-m1 dataset with a forecasting horizon of 96.}
  \label{efficiency}
  \begin{tabular}{c|c|c|c|c|c|l}
    \toprule
    Models & LLM in model? & Total Params. & Trainable Params. & \% Trained & Inference GPU memory & Inference speed\\
    \midrule
    Time-LLM & yes & 135.35M & 53.44M           & 39.49\%             & 895.65 MiB            & 0.0222s/iter\\
TimeCMA  & no & \textbf{18.32M} & 18.32M     & 100.00\%            & 361.93 MiB            & \textbf{0.0094s/iter}\\
BALM-TSF     & yes & 82.88M  & \textbf{0.97M}  & \textbf{1.17\%}     & \textbf{231.30 MiB}   & 0.0218s/iter\\
  \bottomrule
\end{tabular}
\end{table*}

\subsection{Efficiency Analysis}
We compare BALM-TSF with two recent LLM-based baselines, Time-LLM~\cite{jin2023time} and TimeCMA~\cite{liu2024timecma}, using a six-layer GPT-2 on a single NVIDIA A100 GPU. All models were trained on the ETT-m1 dataset with horizon 96 and batch size 8 for fair comparison.

As summarized in Table \ref{efficiency}, BALM-TSF requires only 0.97M trainable parameters (1.17\% of 82.88M total), greatly reducing computation compared to Time-LLM (53.44M) and TimeCMA (18.32M). Notably, TimeCMA relies on a frozen LLM solely to extract embeddings, with no LLM components used during either training or inference, which accounts for its smaller total parameter. During inference stage, BALM-TSF further distinguishes itself with lowest inference GPU memory usage and competitive inference speed. These results demonstrate the lightweight design of BALM-TSF.

\section{Conclusion}

In this work, we tackle the fundamental challenge of modality imbalance in LLM-based time series forecasting. We propose BALM-TSF, a lightweight dual-branch framework that separately processes textual data and raw time series. By introducing a learnable prompt to extract semantic summaries of statistical information and a compact balanced-alignment module that combines horizon-aware scaling with contrastive learning, BALM-TSF effectively mitigates both semantic and distributional imbalances between modalities. This design also successfully leverages the LLM’s pretrained knowledge to enrich forecasting inputs. Extensive experiments demonstrate that our lightweight approach delivers state-of-the-art performance and efficiency in both long-term and few-shot forecasting scenarios.

\begin{acks}
This work is supported by the UoB-Siemens industrial PhD training grant (2226150) and the EPSRC Early Career Researchers International Collaboration Grants (EP/Y002539/1). The computations described in this research were performed using the Baskerville Tier 2 HPC service (https://www.baskerville.ac.uk/). Baskerville was funded by the EPSRC and UKRI through the World Class Labs scheme (EP/T022221/1) and the Digital Research Infrastructure programme (EP/W032244/1) and is operated by Advanced Research Computing at the University of Birmingham. We thank Mengmeng Wang and Chenguang Xiao for their valuable supports.
\end{acks}

\section*{Appendix}
\appendix

\section{Embedding Visualization}
  
\paragraph{Effect of balanced multimodal alignment}

\begin{figure}[ht]
  \centering
  \includegraphics[width=0.5\linewidth]{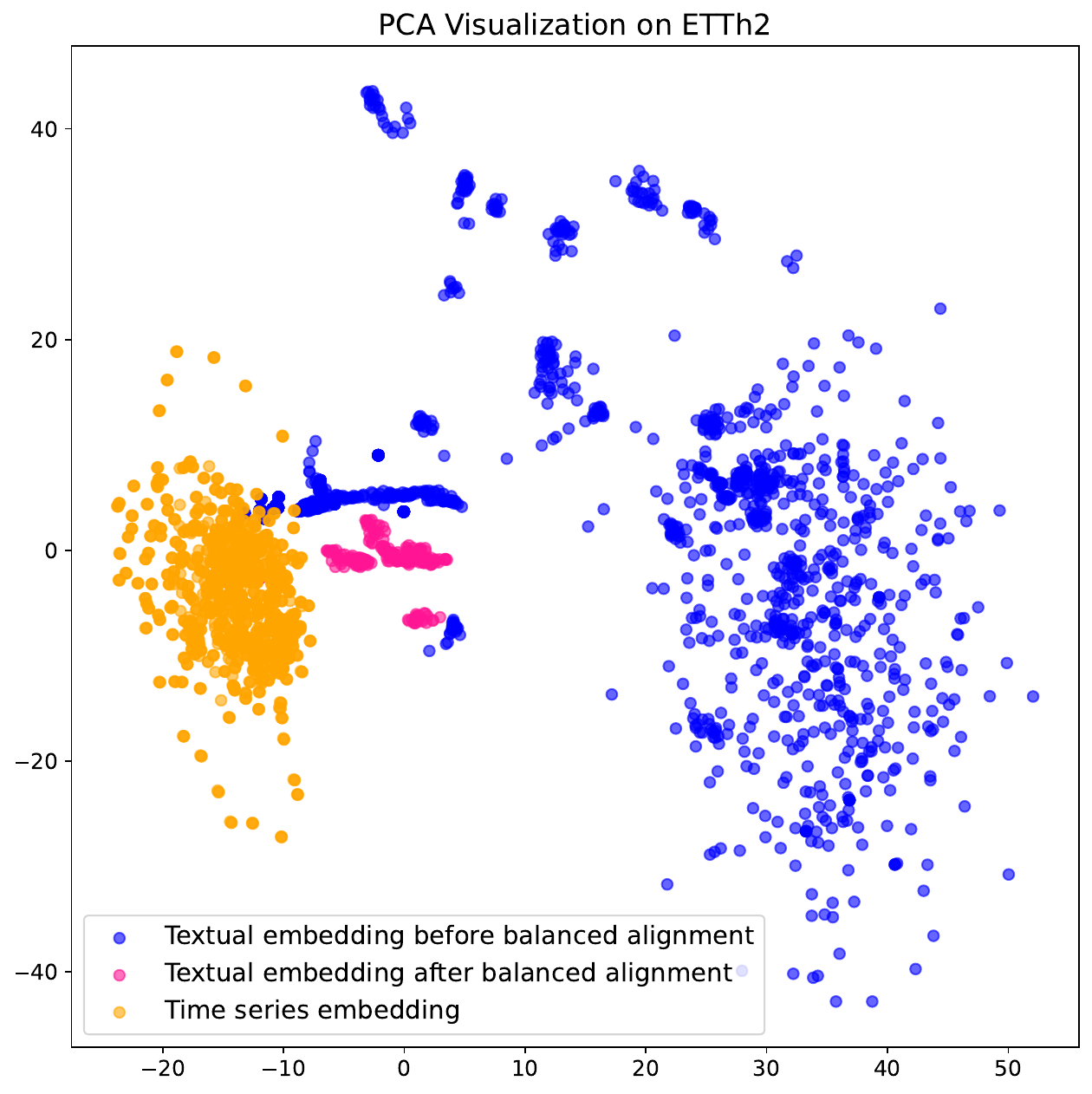}
  \vspace{-1em}
 \caption{PCA visualization on the ETTh2 dataset illustrating the effect of balanced multimodal alignment.}
  \label{fig:pca_ba}
\end{figure}
\begin{figure}[ht]
  \centering
  \includegraphics[width=\linewidth]{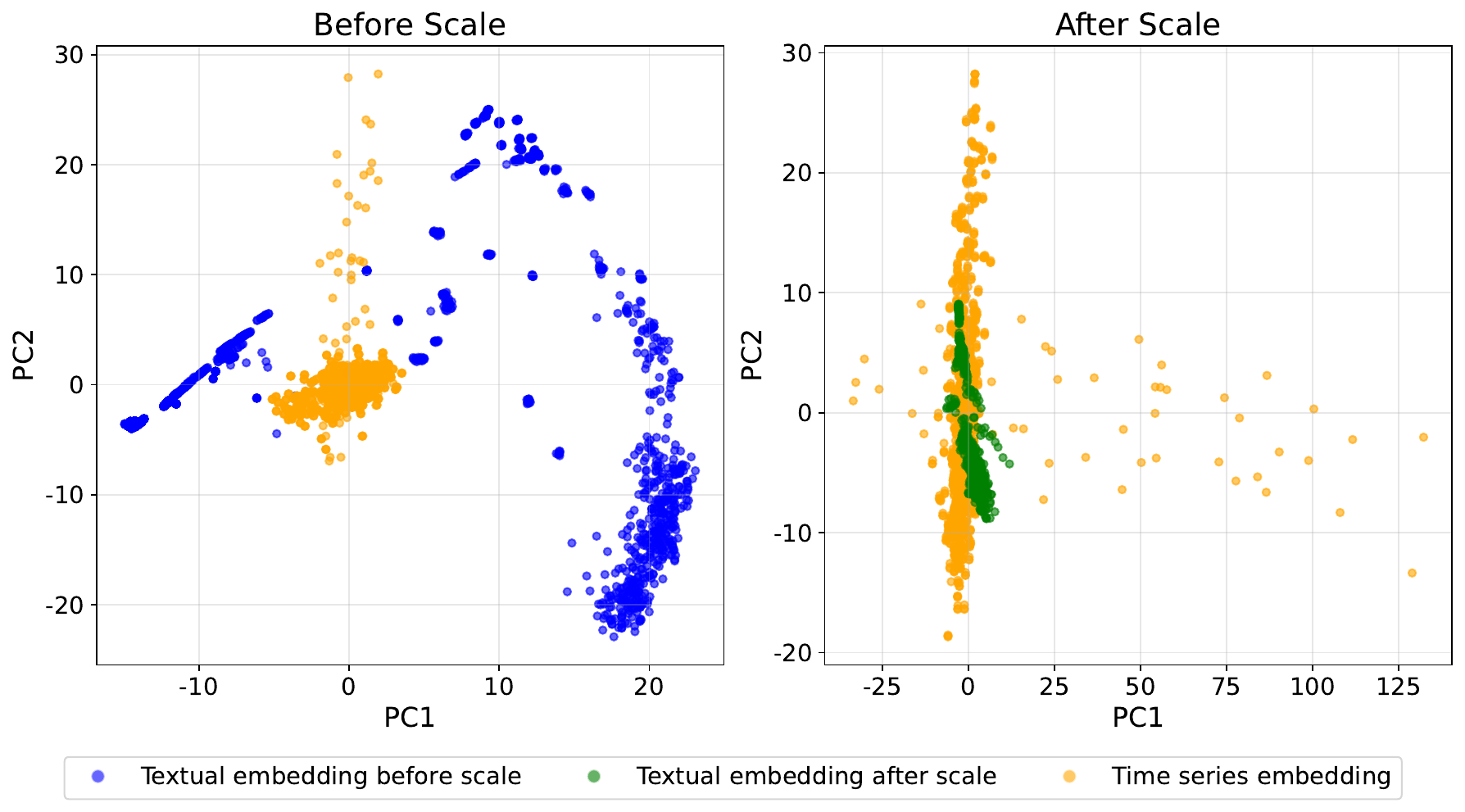}
  \vspace{-2em}
  \caption{PCA visualization on the ETTh2 dataset illustrating the effect of scale.}
  \label{fig:pca_scale}
\end{figure}


We apply Principal Component Analysis (PCA) to BALM-TSF embeddings in Figure~\ref{fig:pca_ba} to visualize the effect of balanced multimodal alignment. The raw textual embeddings (blue) are broadly scattered, reflecting the LLM’s unconstrained encoding, while the time series embeddings (orange) form tighter clusters due to their inherent patterns. After applying balanced multimodal alignment, the adjusted textual embeddings (pink) coalesce into tighter clusters and move closer to the time series embeddings. Since the time series embeddings before and after alignment overlap almost perfectly, we plot them together in orange. This shows that our two-step alignment harmonizes the modalities, enabling BALM-TSF to leverage statistical summaries and enrich time series embeddings for forecasting.

\paragraph{Effect of scale} In Figure~\ref{fig:pca_scale}, the time series embeddings (orange) nearly collapse into a vertical line along PC1, while the original textual embeddings (blue) spread more broadly across PC1/PC2, indicating differences in amplitude and variance. After scaling with standard deviation, the textual embeddings (green) are recentered and rescaled to match the time series distribution, reducing magnitude discrepancies. This strategy thus mitigates distributional imbalance and lays the groundwork for the subsequent alignment.

\section*{GenAI Usage Disclosure}
During paper and code writing, we used LLM to polish the words and code. In the code, we polished loss function with LLM. In the paper, we polished the whole paper with LLM for correct words and clear sentences.

\bibliographystyle{ACM-Reference-Format}
\balance
\bibliography{main}
\end{document}